\RequirePackage{fixltx2e}
\documentclass[11pt,english]{article}
\usepackage{mathptmx}
\usepackage[T1]{fontenc}
\usepackage[utf8]{inputenc}
\usepackage[a4paper]{geometry}
\usepackage{fancyhdr}
\usepackage{color}
\usepackage{babel}
\usepackage{textcomp}
\usepackage{enumitem}
\usepackage{amsmath}
\usepackage{amssymb}
\usepackage{graphicx}
\usepackage{setspace}
\usepackage[authoryear]{natbib}
\usepackage[unicode=true,pdfusetitle,
 bookmarks=true,bookmarksnumbered=false,bookmarksopen=false,
 breaklinks=true,pdfborder={0 0 0},pdfborderstyle={},backref=false,colorlinks=false]
 {hyperref}
\hypersetup{
 linkcolor=blue,citecolor=blue,filecolor=blue,urlcolor=blue}

\makeatletter

\providecommand{\tabularnewline}{\\}

\usepackage{pstricks}
\usepackage{epsfig}
\usepackage{multicol} 
\usepackage{caption}
\usepackage{multirow}
\usepackage{tikz}
\usepackage{qtree}
\usepackage{needspace}

\usepackage{ucs}
\usepackage{tikz-qtree}
\usetikzlibrary{trees}
\usetikzlibrary{calc}

\usepackage{etoolbox}
\usepackage{ragged2e}
\raggedcolumns

\tikzset{square arrow above/.style={to path={-- ++(0,.25) -| (\tikztotarget)}}}
\tikzset{square arrow below/.style={to path={-- ++(0,-.25) -| (\tikztotarget)}}}

\let\OLDthebibliography\thebibliography
\renewcommand\thebibliography[1]{
  \OLDthebibliography{#1}
  \setlength{\parskip}{0pt}
  \setlength{\itemsep}{0pt plus 0.3ex}
}

\let\oldmaketitle\maketitle
\renewcommand{\maketitle}{%
  \oldmaketitle
  \thispagestyle{fancy}
}

\makeatother

\begin{document}
\title{Word length predicts word order: ``Min-max''-ing drives language
evolution}
\author{Hiram Ring\\
NTU Singapore}
\maketitle
\begin{abstract}
A fundamental concern in linguistics has been to understand how languages
change, such as in relation to word order. Since the order of words
in a sentence (i.e. the relative placement of Subject, Object, and
Verb) is readily identifiable in most languages, this has been a productive
field of study for decades (see \citealt{Greenberg:1963ul,Dryer:2007aa,Hawkins:2014aa}).
However, a language’s word order can change over time, with competing
explanations for such changes (\citealt{Carnie:2000fr,Crisma:2009aa,Martins:2018aa,Dunn:2011ul,Jager:2021aa}).
This paper proposes a general universal explanation for word order
change based on a theory of communicative interaction (the Min-Max
theory of language behavior) in which agents seek to minimize effort
while maximizing information. Such an account unifies opposing findings
from language processing (\citealt{Piantadosi:2011aa,Wasow:2022aa,Levy:2008aa})
that make different predictions about how word order should be realized
crosslinguistically. The marriage of both “efficiency” and “surprisal”
approaches under the Min-Max theory is justified with evidence from
a massive dataset of 1,942 language corpora tagged for parts of speech
(\citealt{Ring:2026ab}), in which average lengths of particular word
classes correlates with word order, allowing for prediction of basic
intransitive word order from diverse corpora. The general universal
pressure of word class length in corpora is shown to give a stronger
explanation for word order realization than either genealogical or
areal factors, highlighting the importance of language corpora for
investigating such questions
\end{abstract}

\section{Introduction}

There are two major approaches to studying word order. The first (typological)
approach is to observe word orders in natural languages and identify
patterns whereby the orders of certain constituents are correlated
with other grammatical patterns. This approach has developed multiple
crosslinguistic databases with expert determinations of word order
(such as WALS, \citealt{Dryer:2013ab}; AUTOTYP, \citealt{Bickel:2023aa};
Grambank, \citealt{Skirgard:2023ab}) and has identified statistical
universals in which the orders of certain elements increase the likelihood
for other patterns to be present in a language (\citealt{Greenberg:1963ul,Dryer:2007aa,Dryer:1992fk,Dryer:2011aa,Verkerk:2025aa}).
The second (processing) approach focuses on experimental evidence
for how constituents in particular languages are processed in the
brain. This approach has identified potential universal tendencies
or preferences in processing, notably a preference for “agents first”
(\citealt{Culbertson:2012aa,Futrell:2015ab,Sauppe:2023aa}), “efficiency”
(\citealp{Zipf:1949aa,Wasow:2002aa,Futrell:2015aa}), and “surprisal”
(\citealt{Hale:2001aa,Levy:2008aa}).

While the typological observations do not make predictions about a
language’s basic word order (the statistical universals take basic
word orders as a starting point and make further predictions, see
\citealt{Verkerk:2025aa}), the processing research does imply some
particular claims and predictions about how words should be ordered
in languages of the world. For example, the “agents first” (or “Subject
first”) bias found in experiments suggests that there is a universal
psychological pressure for speakers of all languages to produce agent
arguments (“do-ers” of actions, typically represented by animate nouns
and pronouns) before other kinds of constituents in sentences. This
aligns with observations of the dominant crosslinguistic word order
being Verb-final or Subject-initial (\citealt{Schouwstra:2014aa,Schouwstra:2022aa}).

At the same time, findings from additional language processing research
suggests that other pressures mediate the realization of word order
patterns, namely in relation to “efficiency” and “surprisal”. The
efficiency literature largely follows from Zipf’s (\citeyear{Zipf:1932aa,Zipf:1949aa})
observations that a) languages show a logarithmic distribution whereby
the most frequently used words occur much more often than the next
most frequent, and that b) the most frequent words in a language are
also the shortest. The finding that shorter (or less complex) elements
are more likely to occur initially in a sentence, while longer (or
more complex) elements occur later, is one that can be traced to \citet{Behaghel:1909aa},
and the general idea is that this supports efficient communication.
In contrast to this view, the surprisal literature focuses on how
salient or predictable items are, suggesting that longer (or more
complex) words are more likely to occur first in a sentence since
they improve predictability of following constituents (\citealt{Hale:2001aa,Levy:2008aa,Sauppe:2021aa}).
This is supported by clinical evidence (\citealt{Rezaii:2023aa})
and information theoretic studies (\citealt{Zaslavsky:2020aa,Tucker:2025aa}).

Such studies have implications for particular languages regarding
the ordering of constituents. For example, if a language has longer
verbs than nouns, we would expect the language to be Verb-final based
on the efficiency literature (longer or more complex words last),
whereas the surprisal literature would predict it to be Verb-initial
(longer or more complex words first). These competing predictions
represent hypotheses that have yet to be tested to any great degree
(though some attempts have been made, at least in relation to dependency
structures, see \citealt{Jing:2021aa,Hahn:2022aa}).

The lack of testing highlights the complexity of investigating pressures
that influence the evolution and realization of language structures,
as well as the dearth of datasets and databases that could address
these questions.\footnote{Datasets that have been used by previous research include the Universal
Dependencies Treebanks (UDT; \citealt{Zeman:2024ab}) and ODIN (\citealt{Xia:2016aa})
projects. The UDT contains 160+ hand-annotated languages available
for crosslinguistic investigation, but these are not parallel texts
and the number of sentences in each language is variable, with some
having fewer than 200 sentences in the v2.14 release. The ODIN project
(and its re-packaged form via LECS: https://huggingface.co/datasets/lecslab/glosslm-corpus)
has interlinear glossed text (IGT) from over 1,800 languages, but
the majority of these are single-word/line examples, and only 300
of the languages in their dataset have more than 100 sentences. Additional
cross-linguistic investigations using tagged corpora include \citeauthor{Hahn:2020aa}
(2020: 51 languages), \citeauthor{Jing:2021aa} (2021: 71 languages),
and \citeauthor{Hahn:2020aa} (2022: 80 languages). For each of these
studies the majority of languages are from the Indo-European language
family. Besides limiting generalizability of findings, such data risks
bias toward a particular group of languages (\citealt{Blasi:2022aa}),
in much the same way that other scientific findings have historically
been biased toward WEIRD demographics (\citealt{Henrich:2010aa}).} In particular, questions regarding the length of constituents in
relation to their order in sentences requires a large number of sentences
tagged for parts of speech (at minimum). Given that languages develop
over time via cultural evolution, it is also difficult to separate
general cognitive pressures from historical (genealogical or areal)
factors. This means that inferring crosslinguistic or universal principles
necessitates both a large number of languages as well as sufficient
diversity of languages to establish statistical validity.

\begin{figure}[h]
\noindent \begin{centering}
\includegraphics[width=0.4\columnwidth]{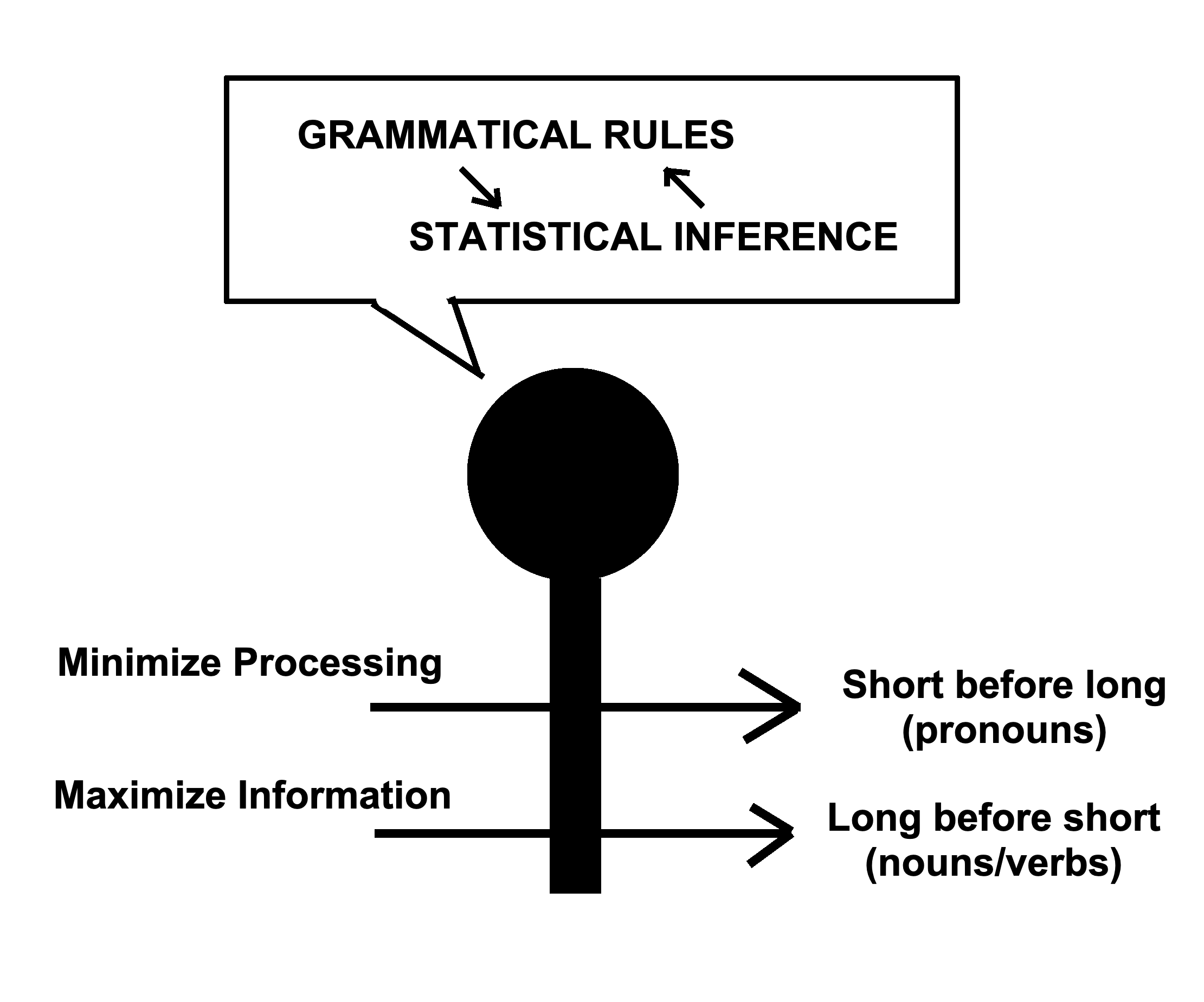}
\par\end{centering}
\caption{Min-Max schema\label{fig:Min-Max-schema}}
\end{figure}

Additionally, the fact that languages develop over time through communicative
interaction requires that any explanation for universal principles
be related to such interactions. That is, in an evolutionary linguistic
paradigm it is individuals (speakers) in a population (language) that
drive variation and select particular linguistic variants for propagation.
The selection process itself is likely motivated by various factors,
two of which are combined in the Min-Max theory of language behavior.
The Min-Max theory proposes that in production speakers select variants
based on two competing pressures: the pressure to minimize effort
and the pressure to maximize information (Figure \ref{fig:Min-Max-schema}).
This language behavior is considered to be core to what drives selection
and propagation of linguistic constructions, and thus change over
time (in, i.e. word order).

The current paper takes advantage of a massive tagged parallel corpus
to show how these pressures are realized in actual language data in
relation to word order. While the dataset is still under development,
the baseline data is tagged for parts of speech, which allows us to
derive a proxy measure for word order from individual language corpora.
It also allows for measurement of other linguistic properties such
as the average length of various word types, while also controlling
for potential genealogical and areal bias. Results show that basic
word order in intransitive clauses is highly correlated with the length
of particular word classes, allowing for the prediction of a language’s
word order based on corpora not in the dataset and explaining variance
better than both genealogy and areal factors.

In the remainder of this paper I first highlight the properties of
the dataset (§\ref{sec:Methodology-and-hypotheses}). I then present
three studies that investigate word length in relation to word order
(§\ref{subsec:Study-1:-Word}), the predictive validity of word lengths
for word order (§\ref{subsec:Study-2:-Predictive}), and the variance
predicted by word length in regression models over and above language
family and linguistic area (§\ref{subsec:Study-3:-Testing}). I conclude
with the implications of these studies in relation to language evolution
(§\ref{sec:Discussion}), discussing how the Min-Max theory of language
behavior unites competing pressures (§\ref{subsec:A-Min-Max-theory}),
as well as some limitations and future directions regarding the current
effort (§\ref{subsec:Limitations}).

\section{Methodology and hypotheses\label{sec:Methodology-and-hypotheses}}

\begin{figure}[h]
\noindent \begin{centering}
\includegraphics[width=0.29\textwidth]{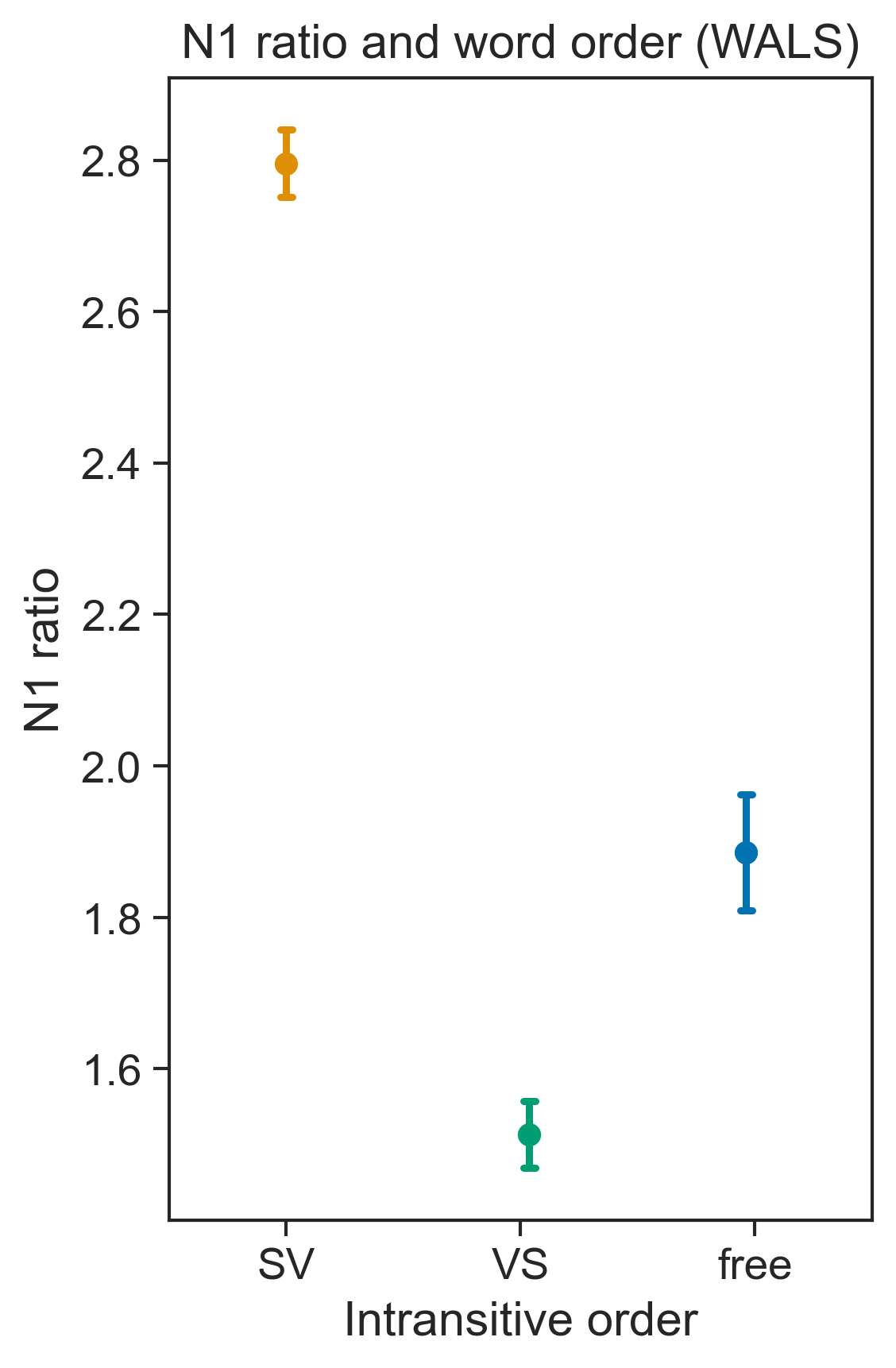}\includegraphics[width=0.3\textwidth]{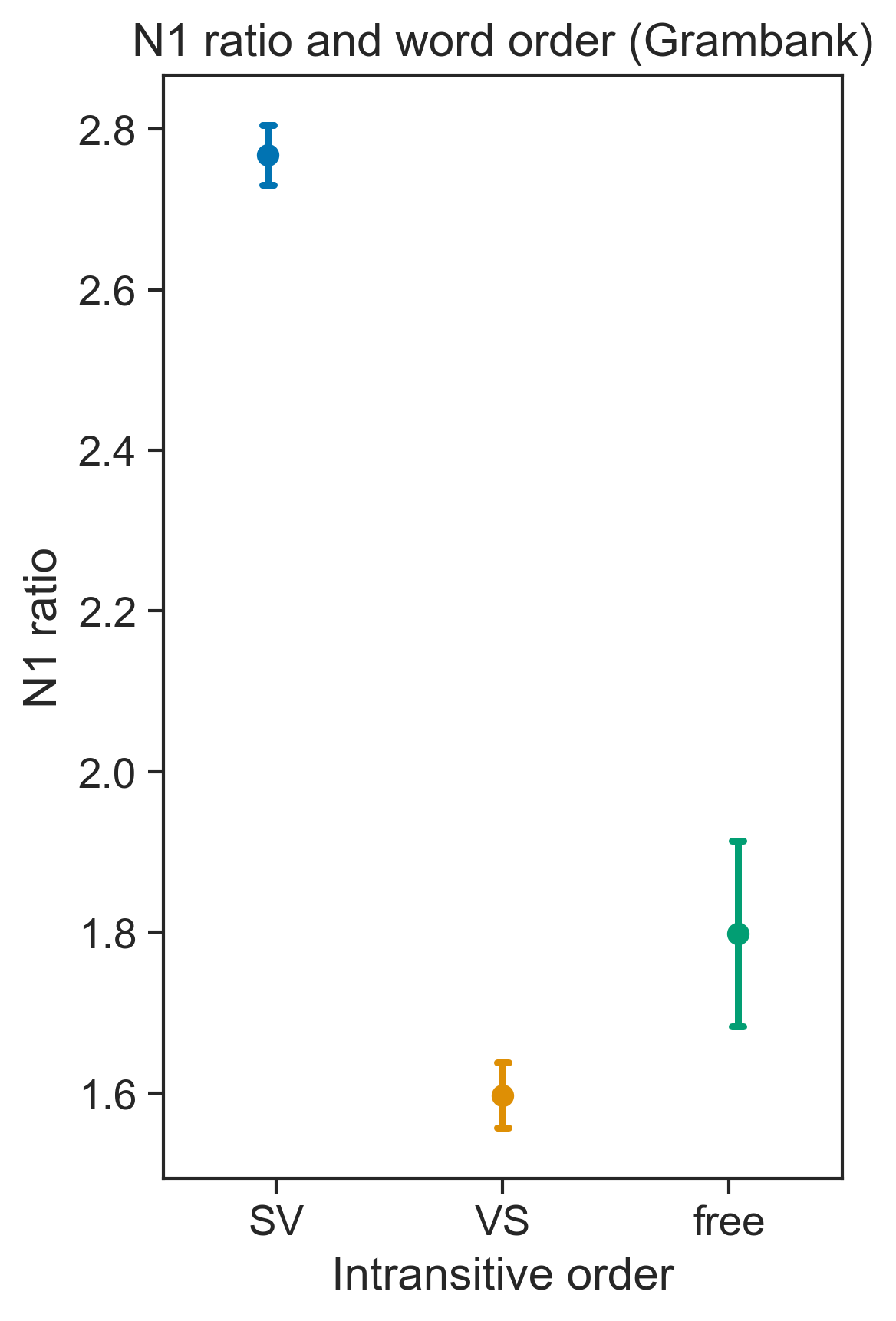}\includegraphics[width=0.305\textwidth]{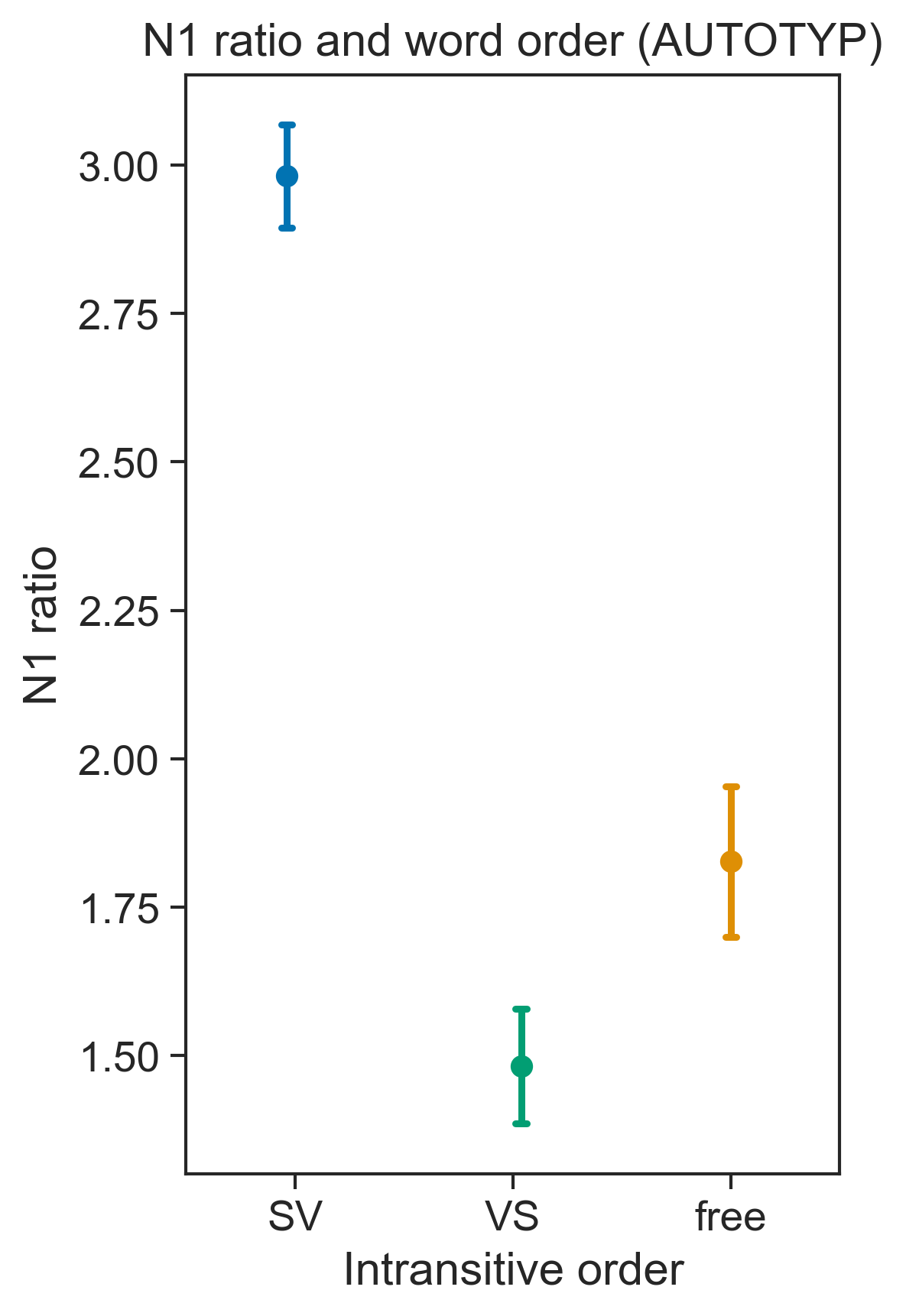}
\par\end{centering}
\caption{The N1 ratio and word order in 3 typological databases (p < 0.001;
\citealt{Ring:2026ab})\label{fig:The-N1-ratio}}
\end{figure}

The \emph{taggedPBC} (\citealt{Ring:2026ab})\footnote{https://github.com/lingdoc/taggedPBC\label{fn:linked-repo} - see
fn \ref{fn:Python-code-and} for a link to a minimal OSF repository
that allows for replication of the studies in this paper.} contains pos-tagged parallel text data in 1,942 languages (roughly
25\% of the world's languages), with over 1,800 parallel verses for
more than 1,850 of those languages. The dataset represents 155 language
families and 78 isolates, dwarfing previously available resources,
and the tagging methodology (via crosslingual tag transfer) has been
shown to correspond well with existing automated taggers for high-resource
languages (as found in the SpaCy and TranKit NLP libraries).\footnote{https://spacy.io ; https://nlp.uoregon.edu/trankit}
Additionally, a novel measure extracted from individual corpora in
this dataset (the ``N1 ratio'') significantly correlates (p < 0.001)
with expert determinations of a language's basic word order for intransitive
clauses (``SV'', ``VS'', ``free''; Figure \ref{fig:The-N1-ratio})
as identified in the typological databases WALS, Grambank, and AUTOTYP.\footnote{The procedure for determing the N1 ratio is essentially to count the
number of sentences in a corpus that have both arguments (nouns, pronouns,
proper nouns) and predicates (verbs, auxiliaries), and then to derive
a ratio from the number of argument-first sentences divided by the
number of predicate-first sentences in that set. As an example, English
{[}ISO 639-3: eng{]} has an N1 ratio of 4.51, whereas Irish {[}ISO
639-3: gle{]} has an N1 ratio of 1.24, based on their respective corpus
in the \emph{taggedPBC}. This ratio can then be related to other aspects
of language, such as word order classifications. See \citet{Ring:2026ab}
for further details.} What this means is that using this dataset we can derive word order
and word length information in a systematic manner, allowing for an
investigation of word length in relation to word order.

The following studies focus on intransitive word order (i.e. the basic
word order of intransitive clauses, in which a predicate requires
a single ``S'' argument) in languages found in the \emph{taggedPBC}.\footnote{Minimal data and Python code to replicate the results in this paper
can be found at https://osf.io/nsaq2 (3MB). More comprehensive data
and code (7GB) can be found at the \emph{taggedPBC} github repository
(fn \ref{fn:linked-repo}).\label{fn:Python-code-and}} The first study (§\ref{subsec:Study-1:-Word}) measures word classes,
showing that average lengths of nouns and verbs in a language's corpus
are highly correlated with SV, VS, and free word order classifications
when frequency is considered, and that this holds even when controlling
for family and areal bias. The second study (§\ref{subsec:Study-2:-Predictive})
investigates whether this effect occurs outside of the \emph{taggedPBC},
showing that information on average lengths of nouns and verbs computed
from different corpora also allow for the classification of languages
as SV or VS, differentiating between historical and modern languages
for which word order changes are known. The third study (§\ref{subsec:Study-3:-Testing})
examines whether the lengths of nouns and verbs in corpora provides
a better indication of word order than known areal or genealogical
signals, and shows that in heirarchical linear regressions, word length
is indeed a better predictor of word order.

\subsection{Study 1: Word length and word order\label{subsec:Study-1:-Word}}

Study 1 follows previous research in using word length as a measure
of word complexity (\citealt{Lewis:2016aa,Bentz:2016aa,Bentz:2017aa,Baerman:2015aa,Baerman:2017aa,Bentz:2023aa};
and see \citealp{Piantadosi:2011aa,Meylan:2021aa,Koplenig:2022aa}
for further discussion in relation to frequency in corpora), comparing
average lengths of particular classes of words (nouns/arguments and
verbs/predicates) with word order in individual languages.\footnote{In the following studies romanized orthographic characters are used
as a proxy for phonemes.} The two processing claims in the literature (from ``efficiency''
and ``surprisal'' approaches) give rise to the following two competing
hypotheses regarding this relationship:
\begin{description}
\item [{H1:}] SV languages have shorter ``S'' than ``V'', while VS
languages have shorter ``V'' than ``S''.\\
\emph{The average length of nouns is shorter in languages classed
as Subject-initial, while the average length of verbs is shorter in
languages classed as Verb-initial.}\textbf{}\\
\textbf{\textcolor{white}{\textvisiblespace{}}}\textbf{\hfill{}(efficiency
- minimize complexity)}
\item [{H2:}] SV languages have longer ``S'' than ``V'', while VS languages
have longer ``V'' than ``S''.\\
\emph{The average length of nouns is longer in languages classed as
Subject-initial, while the average length of verbs is longer in languages
classed as Verb-initial.}\\
\textbf{\textcolor{white}{\textvisiblespace{}}}\textbf{\hfill{}(surprisal
- maximize complexity)}
\end{description}
For the 1,942 languages in the \emph{taggedPBC}, we can identify basic
word order via the N1 ratio. We can also identify the average length
of nouns/arguments and verbs/predicates in each language's corpus
and see how well this correlates with the N1 ratio.

\subsubsection{Word order}

Of the languages in the \emph{taggedPBC}, 1,152 are also found in
typological databases with categorizations for word order. For the
remainder, the N1 ratio was used to impute the (SV, VS, free) category
of the remaining 790 languages in the \emph{taggedPBC} not categorized
by WALS, Grambank or AUTOTYP, and the final categorizations were stored
for all 1,942 languages.\footnote{I trained a Gaussian Naive Bayes classifier on the known data to predict
the category (SV, VS, free) of languages for which word order status
was unknown, based on the N1 ratio as derived from the \emph{taggedPBC}.
Code to reproduce these results can be found at the linked repository
(fn \ref{fn:linked-repo}).} This resulted in 1,548 languages classified as ``SV'', 331 languages
classified as ``VS'', and 63 languages classified as ``free''.

\subsubsection{Measuring word length by class}

To measure word length, I counted the number of characters in a word
tagged as Noun/Argument or Verb/Predicate\footnote{Here I conducted several tests to see if there were different effects
for different word types and groupings, where ``argument'' is a
grouping of words tagged as NOUN, PRON, PROPN in the universal tag
set, while ``predicate'' refers to a grouping of VERB and AUX tags.} and computed an average length for each class on a per-language basis.\footnote{For example, English {[}ISO 639-3: eng{]} arguments have an average
length of 5.9 characters, while English predicates have an average
length of 5.75. English nouns have an average length of 5.9 characters,
while English verbs have an average length of 5.8 characters. Irish
{[}ISO 639-3: gle{]} arguments have an average length of 6.31 characters,
while Irish predicates have an average length of 6.29 characters.
Irish nouns have an average length of 5.2 characters, while Irish
verbs have an average length of 7.9 characters.} In the \emph{taggedPBC} the average number of (unique) arguments
per language is 741, and the average number of predicates is 498.
The mean length of arguments is 6.71 characters, and the mean length
of predicates is 6.82 characters. Conducting a Student's T-Test showed
that this is statistically significant (p < 0.001), indicating that
crosslinguistically, arguments are shorter than predicates. All statistics
showed a normal distribution.

Since we are working with corpus data, there are two different length
measures that we can compute for word classes in each language. We
can compute average lengths for classes based on 1) the unique items
in that class, and 2) the total items in the class. The first measure
is similar to the information that can be gleaned from a wordlist
or dictionary, where infrequent terms have the same weight as frequent
items. The second measure gives more weight to frequent items. Since
we know that frequency of occurrence in corpora is a factor in constituent
predictability (\citealp{Piantadosi:2011aa,Meylan:2021aa,Koplenig:2022aa}),
it is worth observing whether the two measures differ in terms of
their relationship with word order.

\subsubsection{Relationship between word order and noun/verb lengths}

\begin{figure}[h]
\noindent \includegraphics[width=0.48\columnwidth]{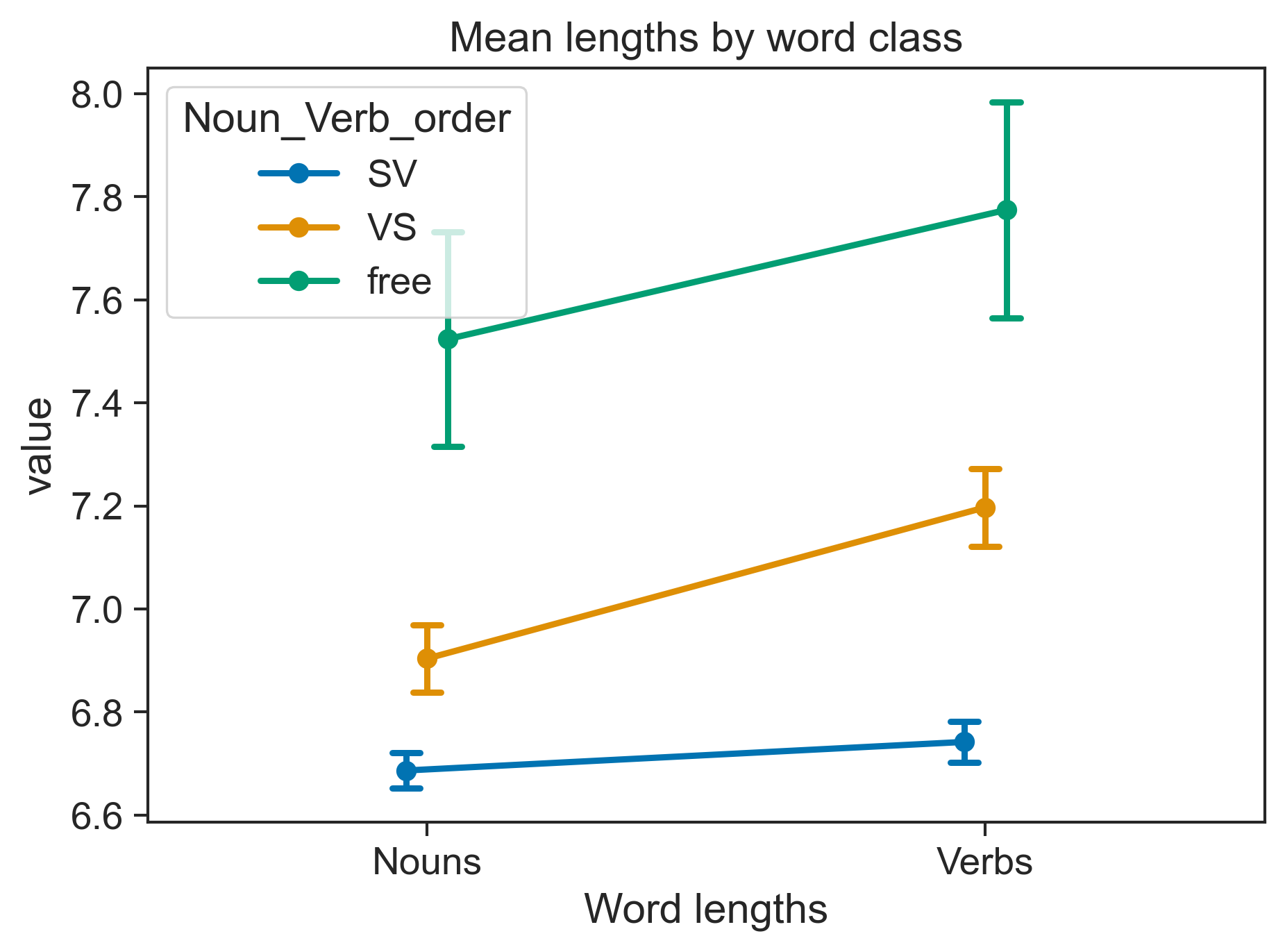}

\caption{Normalized lengths of Nouns, Verbs, by word order\label{fig:Normalized-lengths-of}}

\end{figure}

\begin{figure}[h]
\noindent \includegraphics[width=0.48\columnwidth]{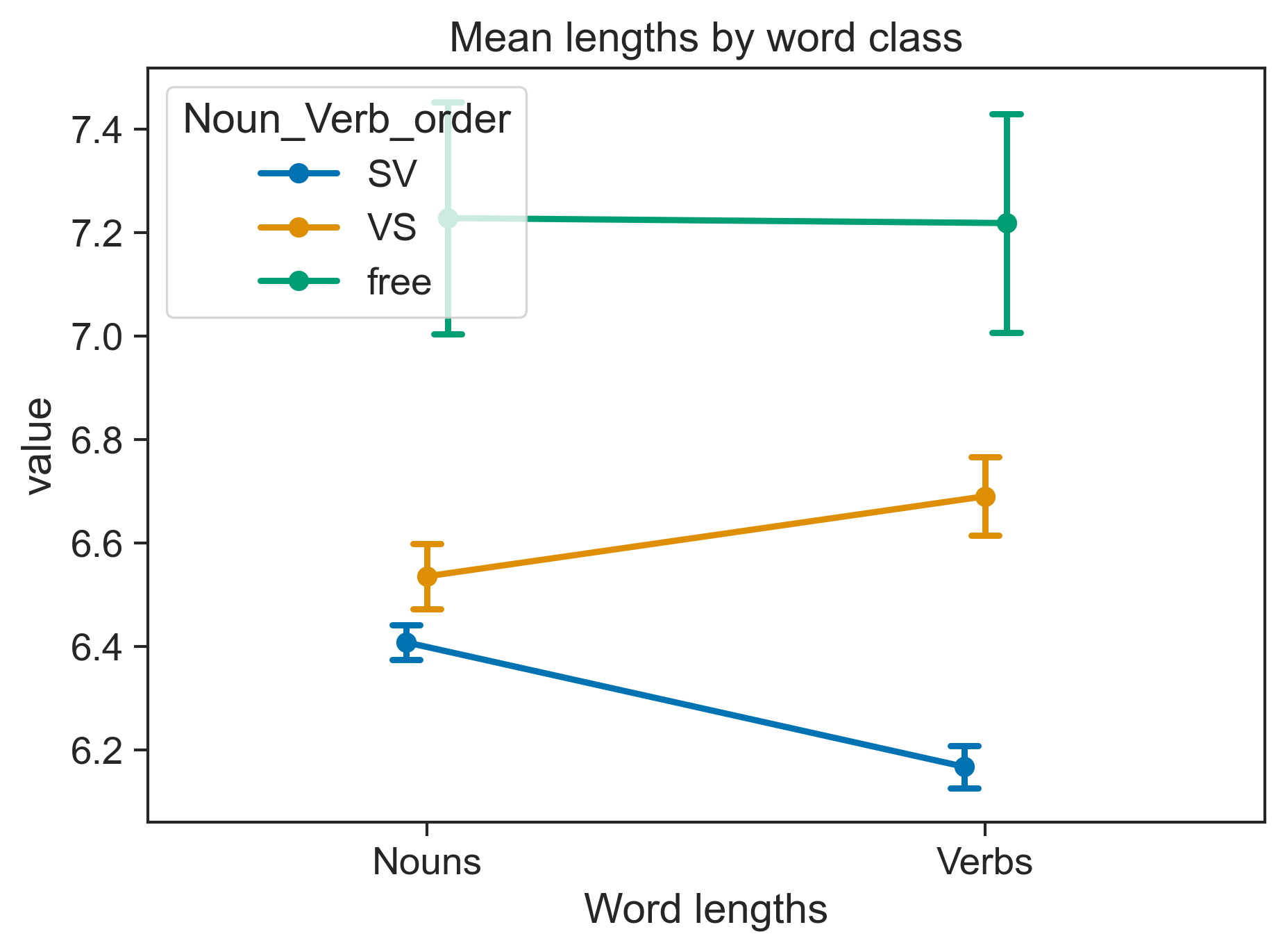}

\caption{Frequency-weighted lengths of Ns, Vs, by word order\label{fig:Frequency-weighted-lengths-of}}

\end{figure}

To see whether word order is related to the length of nouns and verbs
we can first visualize the relationships by plotting the means of
word class lengths (for arguments, predicates, nouns, verbs, etc)
in relation to word order (SV, VS, free). Figure \ref{fig:Normalized-lengths-of}
shows the means of unique nouns and verbs (measure \#1 above) in relation
to word order, while Figure \ref{fig:Frequency-weighted-lengths-of}
shows the means of frequency-weighted nouns and verbs in relation
to word order (measure \#2 above).\footnote{Additional statistical comparisons and plots are available at the
linked repository (Fn \ref{fn:linked-repo}).} Observationally, the plots indicate that when frequency is taken
into account, nouns are longer than verbs in SV languages but shorter
than verbs in VS languages.\footnote{It is also worth noting here that in this corpus words are overall
shorter in SV languages than VS languages, which in turn are shorter
than those in `free' languages. This displays a general observed trend
that languages with free word order tend to have more morphology to
identify argument roles and other grammatical relations (\citealt{Koplenig:2017aa,Yadav:2020aa}).}

To verify this I conducted a repeated-measures ANOVA with the average
lengths of nouns and verbs as the repeated measures and the word order
(SV, VS, free) as the between subject factors. Results of the ANOVA
indicate that the effect of frequency-weighted noun and verb lengths
is highly significant (p < 0.001) for differentiating between word
orders. However, the ANOVA does not account for potential effects
of language relatedness or geographical area, a subject we turn to
next.

\subsubsection{Controlling for genealogical and areal bias\label{subsec:Controlling-for-genealogical}}

Given that human languages develop over time through interaction,
it has long been recognized that the evolution of particular linguistic
features can potentially be accounted for by descent from a common
ancestral state or by contact between languages spoken in the same
area (i.e. Galton's problem, c.f. \citealt{Bromham:2024aa}). It is
therefore common practice to account or control for such genealogical
or areal effects in typological studies. While there are several proposals
on the best way to handle this (ibid, see also extensive discussion
in \citealt{Guzman-Naranjo:2022aa}), a basic method is to group observed
languages according to their family membership and/or macroarea.

In keeping with such approaches, for each language I identified family
membership and linguistic area on the basis of Glottolog classifications,\footnote{Glottolog (\citealt{Hammarstrom:2024aa}; https://www.glottolog.org)
contains expert annotations for language family affiliations for over
7,000 languages, along with `macroarea' identifiers and additional
aggregated data from a variety of sources. For family membership I
take the highest level node, based on the observation that all modern
languages are the same age, and while their pathways/histories may
differ, this does not impact whether they can be considered part of
the same phylogenetic tree.} filling in missing data via Ethnologue and other sources, and included
them as grouping criteria in a linear mixed-effects regression model.
From the dataset of 1,942 languages I examined four word-length-related
continuous features to identify potential effects of genealogy and
area on word order. The first two features are the frequency-weighted
average length of nouns and verbs in a corpus. The third feature is
the difference in length between nouns and verbs, which is related
to minimization of complexity, based on the idea that a larger difference
between lengths of word classes may effect the likelihood of that
particular class being placed first. The fourth feature is the ratio
of noun and verb lengths, and is related to maximization of complexity
(via ``entropy''). Here a lower ratio indicates that the particular
word classes are less likely to be easily differentiable (and therefore
predictable) based on their length.

For each feature I conducted separate linear regressions, treating
both grouping factors (family membership, macroarea) as random effects
with word order (SV, VS, free) as the dependent variable and the feature
as the independent variable (Table \ref{tab:Word-length-features-1}).\footnote{Tables in this section use the Mixed Linear Model from the Python
\emph{statsmodels} library, and all features are scaled with the MinMaxScaler
from \emph{scikit-learn} (which scales values to between 0 and 1).
Other approaches can easily be explored using code at the \emph{taggedPBC}
Github repository.} Each feature was found to be significant, with little effect from
either family membership or macroarea. Treating family and macroarea
as nested within each other did not change this result.

\begin{table}[h]
\begin{centering}
\begin{tabular}{lllll}
\multicolumn{5}{l}{Number of Languages: 1,942}\tabularnewline
iv & Coef. & Std.Err. & z & P>|z|\tabularnewline
\hline 
 &  &  &  & \tabularnewline
NVlensdiff & 0.281 & 0.094 & 2.996 & 0.003\tabularnewline
NVlensratio & 0.255 & 0.080 & 3.174 & 0.002\tabularnewline
Nlen\_freq & 0.433 & 0.109 & 3.957 & 0.000\tabularnewline
Vlen\_freq & 0.499 & 0.100 & 4.998 & 0.000\tabularnewline
\end{tabular}
\par\end{centering}
\caption{Word length features in relation to word order (SV, VS, free) with
family/area factors (for 1,942 languages)\label{tab:Word-length-features-1}}

\end{table}

As an additional check (Table \ref{tab:Word-length-features-2}),
I filtered the dataset based on number of languages in each family,
first excluding single languages and isolates (leaving 1,864 languages),
then excluding families with fewer than 10 members (leaving 1,653
languages). In each case all features remained significant, with little
effect from family membership or macroarea.

\begin{table}[h]
\begin{centering}
\begin{tabular}{lllll}
\multicolumn{5}{l}{Number of Languages: 1,864}\tabularnewline
iv & Coef. & Std.Err. & z & P>|z|\tabularnewline
\hline 
 &  &  &  & \tabularnewline
NVlensdiff & 0.321 & 0.095 & 3.360 & 0.001\tabularnewline
NVlensratio & 0.283 & 0.081 & 3.478 & 0.001\tabularnewline
Nlen\_freq & 0.448 & 0.111 & 4.045 & 0.000\tabularnewline
Vlen\_freq & 0.539 & 0.101 & 5.318 & 0.000\tabularnewline
\end{tabular}\medskip{}
\par\end{centering}
\begin{centering}
\begin{tabular}{lllll}
\multicolumn{5}{l}{Number of Languages: 1,653}\tabularnewline
iv & Coef. & Std.Err. & z & P>|z|\tabularnewline
\hline 
 &  &  &  & \tabularnewline
NVlensdiff & 0.382 & 0.101 & 3.766 & 0.000\tabularnewline
NVlensratio & 0.323 & 0.086 & 3.751 & 0.000\tabularnewline
Nlen\_freq & 0.332 & 0.122 & 2.710 & 0.007\tabularnewline
Vlen\_freq & 0.475 & 0.108 & 4.394 & 0.000\tabularnewline
\end{tabular}
\par\end{centering}
\caption{Word length features in relation to word order (SV, VS, free) with
family/area factors (for 1,864 and 1,653 languages respectively)\label{tab:Word-length-features-2}}
\end{table}

It is worth noting here that coefficients of the independent variables
are in most cases not above 0.5. However, when we see that these coefficients
are combined with z-values above 2.0 and highly significant P-values
(all below 0.01), it indicates a robust effect (\citealt{Luke:2017aa})
for each feature. It is also important to observe that two of the
features are derived from the others, which means there is likely
to be high collinearity, such that combining their values in a regression
results in the most robust feature ('Vlen\_freq' in this case) being
found to be significant while the others are not. That features are
found to be collinear does not mean that they should be excluded from
a model, however - features that point in the same direction can still
add extra information, which can be useful in other tasks such as
classification, to which we turn next. These observations from linear
regressions are simply one way of looking at the data, highlighting
the importance of word length in relation to word order. By including
family and macroarea as independent random effects the results indicate
that this relationship is not biased by either factor.

\subsection{Study 2: Predictive validity of word length\label{subsec:Study-2:-Predictive}}

Given the observation that relative lengths of nouns and verbs differ
in corpora for languages with different word orders, we can ask how
predictive this observation is. That is, can we predict the word order
of a language based on the relative lengths of its nouns and verbs?
If we are strictly interested in identifying basic word order from
corpora, it is clear that the N1 ratio already allows us to do so.
Study 1 has shown that word length features derived from corpora are
also highly significant for differentiating word orders. This is relevant
because we are interested in understanding how noun and verb lengths
may impact how word order changes over time. Given two sets of corpora
from two timepoints of a language, being able to predict word order
from noun and verb lengths in a corpus for both the modern language
and its ancestral language would indicate that such properties are
related to how word order evolves.

To investigate whether this is the case, for Study 2 I sourced additional
corpora of tagged data from two pairs of historical and modern languages
in which the the descendent changed its word order from the ancestral
state:
\begin{enumerate}
\item Ancient Hebrew (VS) and Modern Hebrew (SV)
\item Classical Arabic (VS) and Egyptian Arabic (SV)
\end{enumerate}
All languages belong to the Semitic branch of the Afroasiatic language
family, a necessary constraint given the limited amount of data available
for most pre-modern languages. In the case of Hebrew, the UDT has
POS-tagged sentences for both Ancient\footnote{https://universaldependencies.org/treebanks/hbo\_ptnk}
and Modern\footnote{https://universaldependencies.org/treebanks/he\_htb}
varieties. In the case of Classical Arabic I used the Quranic Arabic
Corpus\footnote{Version 0.4: https://corpus.quran.com} and for Egyptian
Arabic I used the BOLT Egyptian Arabic Treebank (\citealt{Maamouri:2018aa}),\footnote{https://catalog.ldc.upenn.edu/LDC2018T23 (Discussion Forum)}
which are also hand-tagged for parts of speech. If the lengths of
nouns and verbs in these corpora allows us to correctly classify Ancient
Hebrew and Classical Arabic as VS and the modern varieties as SV,
this would suggest that these are salient properties affecting how
word order has developed diachronically. Given the textual differences
between the \emph{taggedPBC} (New Testament texts) and the test corpora
(Biblical, Quranic, modern, and informal texts), such a finding would
be quite striking.

In order to conduct this test, the data for Ancient Hebrew (ISO 639-3:
hbo), Modern Hebrew (ISO 639-3: heb), Classical Arabic (cla)\footnote{Classical Arabic does not have an assigned ISO 639-3 code or Glottocode.
Accordingly, I use the abbreviation \emph{cla} here, as there is no
current language assigned to this code in the ISO 639-3 tables.} and Egyptian Arabic (ISO 639-3: arz) were processed and the respective
word classes were counted and measured as per the method outlined
for Study 1 (each corpus > 400 sentences, see Table \ref{tab:Corpora-used-for}).

\begin{table}[h]
\begin{tabular}{llllll}
\textbf{Language} & \textbf{ISO 639-3} & \textbf{Data Source} & \textbf{Sentences} & \textbf{Unique nouns} & \textbf{Unique verbs}\tabularnewline
Ancient Hebrew & hbo & UDT v2.14 & 409 & 734 & 782\tabularnewline
Modern Hebrew & heb & UDT v2.14 & 882 & 2068 & 1297\tabularnewline
Classical Arabic & cla & Quran & 6,236 & 8,243 & 7,966\tabularnewline
Egyptian Arabic & arz & BOLT & 31,688 & 16,698 & 10,581\tabularnewline
\end{tabular}

\caption{Corpora used for testing historical word order change\label{tab:Corpora-used-for}}
\end{table}

Five features related to word length were used to train a Gaussian
Naive Bayes (GNB) classifier, namely the four features described in
§\ref{subsec:Controlling-for-genealogical} above as well as a categorical
feature: whether a language's nouns are longer than its verbs. Using
these five features (nounlen, verblen, nlen-vlen, nlen/vlen, nlonger),
a GNB classifier was trained on the original combined word order dataset
from the \emph{taggedPBC} containing hand-coded classification of
1,150 languages (Modern Hebrew and Modern Arabic were removed).\footnote{As noted previously, since we are concerned here with classification,
the collinearity of relevant features is not an issue.} The trained classifier was then used to predict word order for our
pairs of languages: Ancient Hebrew and Modern Hebrew on the one hand,
Classical Arabic and Egyptian Arabic on the other. Ancient Hebrew
was classified as VS while Modern Hebrew was classified as SV by the
trained classifier. Classical Arabic was classified as VS while Egyptian
Arabic was classified as SV.

To double check this finding, I sourced data from another historical
language, Old Irish, as represented in the Codex Paulinus Wirziburgensis,
whereby Latin text of St. Paul's epistles was annotated by Irish scholars
during the middle of the eighth century. The dataset (Table \ref{tab:Irish-corpus-used};
\citealt{Doyle:2018aa}) contains over 3,600 Irish glosses/sentences.
As an Indo-European language, Old Irish is widely surmised to have
changed its word order to verb-initial from an earlier state (c.f.
\citealt{Watkins:1963aa,Hickey:2002aa}), so this check allows us
to verify whether the same properties hold in relation to lengths
of nouns and verbs. I then removed Old Irish from the training dataset,
retrained the classifier, and predicted on the features of the Würzburg
Glosses. Indeed, the classifier accurately classified Old Irish as
VS, based on the data extracted from the Würzburg Glosses.

\begin{table}[h]
\begin{tabular}{llllll}
\textbf{Language} & \textbf{ISO 639-3} & \textbf{Data Source} & \textbf{Sentences} & \textbf{Unique nouns} & \textbf{Unique verbs}\tabularnewline
Old Irish & sga & Würzburg & 3,647 & 743 & 435\tabularnewline
\end{tabular}

\caption{Irish corpus used for testing historical word order change\label{tab:Irish-corpus-used}}
\end{table}

Study 2 thus indicates that the frequency-weighted relative lengths
of nouns and verbs can predict word order from corpora unrelated to
the type of corpora contained in the \emph{taggedPBC}, with various
implications for our understanding of language processing and language
change. This is a notable finding, and I am unaware of any other research
that has been able to predict word order change based on measurement
of corpus-based features. However, it is possible that this result
is conditioned to some degree by the fact that the majority of the
test cases are Semitic languages spoken in the same geographical area,
so we now turn to an investigation of whether genealogical descent
or location provides a better explanation for the correlation of noun
and verb lengths with word order.

\subsection{Study 3: Testing variance predicted by word length\label{subsec:Study-3:-Testing}}

The preceding studies have shown, first, that features related to
word length differentiate between intransitive word orders crosslinguistically
(Study 1) and, second, that such features can predict intransitive
word order from corpora for both historical and modern languages (Study
2). While Study 1 controlled for the effect of language family and
macroarea on word length features, Study 3 asks a slightly different
question. That is: of the factors known to explain word order patterns
in languages (family, area, word length), which is more explanatory?

\citet{Dunn:2011ul} showed that descent from a common ancestor was
a better predictor of many grammatical patterns than other explanations.
Although \citet{Jager:2021aa} counter this claim with additional
data, we can also use the current dataset to see if descent is more
explanatory of word order for a given language than the lengths of
nouns and verbs. Regarding language area, \citet{Guzman-Naranjo:2022aa}
show that areal affects are as important as family membership in predicting
linguistic phenomena, and that the results in \citeauthor{Dunn:2011ul}
can be just as easily predicted by language area. This is also a claim
that we can test with the \emph{taggedPBC} dataset.

To implement the comparison and identify the relative importance of
known predictors we can use hierarchical linear regression (HLR).
This approach allows researchers to observe the relative impact of
particular features that are theoretically relevant to the question
at hand by seeing how they influence the performance of successive
models. In the case of word order, we can train a base regression
model with a feature known to be relevant to word order (the N1 ratio),
and then subsequent models that each incorporate additional features
of interest (``language area/location'', ``family membership'',
``noun/verb lengths''). We can then compare the models to see whether
the additional features increase the variance explained in the data
via the F-value: an increase in F-value (change) and the P-value of
this change indicates more/additional variance explained by the particular
feature added in a model. Importantly, because we are not concerned
with the significance of included features, each individual model
can contain multiple (potentially collinear) features related to a
given theoretical construct, as what is being measured is simply degree
and significance of the \emph{change} in fit of each model relative
to the previous one.

This approach also means we can run multiple analyses with differing
sets of data in order to see how this affects the fit of the models.
It may be, for example, that using a smaller sample gives a different
result than a larger or more diverse one. Here, the more languages
we have for a language family or linguistic area, the higher the likelihood
that any relationship identified is not due to chance. One caveat
of this approach is that when the dependent variable is categorical,
an HLR requires that it be transformed to a single distinction. Thus
in the ensuing observations I remove languages with ``free'' word
order from consideration, as this leaves us with a binary distinction
in word order (SV, VS) compatible with the HLR approach.

Accordingly, for all languages in the \emph{taggedPBC} I identified
family membership and macroarea as noted in §\ref{subsec:Controlling-for-genealogical}.
Then, after removing the 63 languages with ``free'' word order,
I filtered the data on four conditions: a) languages found in \citeauthor{Dunn:2011ul}
that are also present in the \emph{taggedPBC} (120 languages), b)
languages found in the \emph{taggedPBC} that are in the four families
investigated by \citeauthor{Dunn:2011ul} (662 languages), c) languages
in families with more than 75 members in the \emph{taggedPBC} (1,182
languages), and d) languages in families with 2 or more members in
the \emph{taggedPBC} (1,805 languages). For each of these conditions,
I ran an HLR, adding features of interest progressively to see how
they impacted the fit of the successive models and whether particular
features account for additional variance.

The base model (Model 1) uses the N1 ratio as the primary feature
for classification. This is because we know from comparison with expert
determinations of word order (see \citealt{Ring:2026ab}) that the
N1 ratio significantly correlates with the classification of languages
as SV or VS. Due to the strong relationship observed between language
area and typological properties of language (see \citealt{Guzman-Naranjo:2022aa}),
Model 2 then includes macroarea identifiers for each language as features.
Model 3 additionally incorporates descent (language family membership)
as a feature (per \citealt{Dunn:2011ul}), and Model 4 includes the
average lengths of nouns and verbs (per the relationship identified
in this paper).\footnote{In these models, continuous features (N1 ratio, N/V len) were scaled
using the MinMaxScaler from \emph{scikit-learn}, and categorical features
(family, macroarea) were converted to numeric values.} If descent is more explanatory than the lengths of nouns/verbs, this
should be clear from the differences between the respective models
for each condition. Results are displayed in Table \ref{tab:Hierarchical-Linear-Regressions}.

\begin{table}[h]
\begin{tabular}{clrrrrrl}
a) & Model & N (obs) & F-val & P-val (F) & F-val change & P-val (F-val change) & \tabularnewline
\cline{2-8} \cline{3-8} \cline{4-8} \cline{5-8} \cline{6-8} \cline{7-8} \cline{8-8} 
 & 1 (N1 ratio) & 120 & 45.28 & 6.5e-10 &  &  & \tabularnewline
 & 2 (Lg area) & 120 & 46.63 & 1.28e-15 & 34.95 & 3.41e-08 & \textbf{{*}{*}}\tabularnewline
 & 3 (Family) & 120 & 30.86 & 9.65e-15 & 0.07 & 0.79 & \tabularnewline
 & 4 (N/V len) & 120 & 19.57 & 4.92e-14 & 1.91 & 0.15 & \tabularnewline
\end{tabular}\medskip{}

\begin{tabular}{clrrrrrl}
b) & Model & N (obs) & F-val & P-val (F) & F-val change & P-val (F-val change) & \tabularnewline
\cline{2-8} \cline{3-8} \cline{4-8} \cline{5-8} \cline{6-8} \cline{7-8} \cline{8-8} 
 & 1 (N1 ratio) & 662 & 190.0 & 3.61e-38 &  &  & \tabularnewline
 & 2 (Lg area) & 662 & 182.17 & 1.05e-63 & 135.59 & 1.25e-28 & \textbf{{*}{*}}\tabularnewline
 & 3 (Family) & 662 & 122.02 & 7.73e-63 & 1.47 & 0.23 & \tabularnewline
 & 4 (N/V len) & 662 & 82.34 & 4.44e-67 & 15.02 & 4.18e-07 & \textbf{{*}{*}}\tabularnewline
\end{tabular}\medskip{}

\begin{tabular}{clrrrrrr}
c) & Model & N (obs) & F-val & P-val (F) & F-val change & P-val (F-val change) & \tabularnewline
\cline{2-8} \cline{3-8} \cline{4-8} \cline{5-8} \cline{6-8} \cline{7-8} \cline{8-8} 
 & 1 (N1 ratio) & 1182 & 378.83 & 2.14e-73 &  &  & \tabularnewline
 & 2 (Lg area) & 1182 & 327.01 & 1.05e-113 & 208.55 & 1.2e-43 & \textbf{{*}{*}}\tabularnewline
 & 3 (Family) & 1182 & 218.01 & 1.79e-112 & 0.37 & 0.54 & \tabularnewline
 & 4 (N/V len) & 1182 & 132.79 & 1.16e-111 & 3.55 & 0.03 & \textbf{{*}}\tabularnewline
\end{tabular}\medskip{}

\begin{tabular}{clrrrrrl}
d) & Model & N (obs) & F-val & P-val (F) & F-val change & P-val (F-val change) & \tabularnewline
\cline{2-8} \cline{3-8} \cline{4-8} \cline{5-8} \cline{6-8} \cline{7-8} \cline{8-8} 
 & 1 (N1 ratio) & 1805 & 535.42 & 6.17e-104 &  &  & \tabularnewline
 & 2 (Lg area) & 1805 & 327.44 & 4.99e-122 & 92.34 & 2.35e-21 & \textbf{{*}{*}}\tabularnewline
 & 3 (Family) & 1805 & 218.53 & 6.92e-121 & 0.78 & 0.38 & \tabularnewline
 & 4 (N/V len) & 1805 & 136.84 & 3.62e-123 & 10.76 & 2.27e-05 & \textbf{{*}{*}}\tabularnewline
\end{tabular}

\caption{Hierarchical Linear Regressions under 4 conditions\label{tab:Hierarchical-Linear-Regressions}}
\end{table}

From these results we can see that, as expected, the N1 ratio significantly
predicts word order under all conditions. Additionally, language area
is a significant predictor of word order that adds to the strength
of the classification. Language family (descent) does not add additional
predictive power to our model under most conditions (over language
area), whereas the length of nouns and verbs does. This analysis indicates
that the length of nouns and verbs in a language accounts for more
variance and at the same time is more significant than that language's
family membership in determining its word order classification, under
all test conditions.

When we remove language area from the linear regression,\footnote{See additional statistical results at the linked repositories (Fn
\ref{fn:linked-repo}; Fn \ref{fn:Python-code-and}).} we find that family becomes a significant predictor of word order
for a smaller sample of languages, but that under these conditions
the length of nouns and verbs still accounts for additional variance
and is also highly significant. Under conditions (c) and (d), with
larger and more diverses samples, the model incorporating language
family provides little explanation and is non-significant, while the
subsequent model incorporating length of nouns and verbs explains
additional variance that is highly significant.

This largely aligns with Jäger and Wahl's (\citeyear{Jager:2021aa})
finding regarding the (lower) correlation between descent and linguistic
structures in comparison to other explanations, with the caveat that
Jäger and Wahl included language isolates, which is likely to greatly
diminish the affect of descent. If a large number of languages in
a sample are not related to others, clearly descent can have only
a minimal impact on word order. By including only language families
with two or more members in the taggedPBC (which excludes 74 languages
classified as having either SV/VS order), I show that descent does
not explain much variance above and beyond language location/area,
in contrast with noun and verb lengths. The test where language area
was removed altogether from the HLR models also suggests that language
family/descent is highly correlated with linguistic area, which conforms
with expectations related to human history and migration. Additionally,
the amount of variance explained seems to depend on the number of
languages (and language families) in the sample, highlighting the
need for more data for languages from under-represented language families.

\section{Discussion\label{sec:Discussion}}

The preceding investigations have shown that the average length of
nouns and verbs, weighted by their frequency in a corpus, correlates
with intransitive word order crosslinguistically. This correlation
is strong enough that such word length features can predict word order
for both historical and modern languages, and is a stronger signal
for word order than either language family or language area. These
findings have various implications for our understanding of language
processing and how languages change under an agentic evolutionary
model. In particular, the effect of frequency and its relation to
processing suggests some possibilities for reconceptualizing a theory
of language behavior.

The fact that a clear relationship between word order and word length
is not observed unless frequency is taken into account indicates that
the availability of corpus data is required to investigate questions
related to crosslinguistic grammatical patterns and how they might
develop. Here, the ability to group classes of words for observation
allows for more fine-grained analyses regarding such patterns. For
example, when we group word classes into arguments (prototypical agents/actors
as represented by nouns, pronouns, proper nouns) and predicates (prototypical
actions/states as represented by verbs, auxiliaries) there is a different
realization of length than if we solely consider nouns and verbs.
In the former condition all arguments are shorter than verbs crosslinguistically,
while in the latter condition the differences in length emerge.

That arguments overall are shorter in languages of the \emph{taggedPBC},
across all word orders, suggests that there are certain (non-Noun)
arguments affecting these observations. While this has yet to be investigated
in detail, the effect of PRON seems quite likely, whereby a tendency
for pronouns to be shorter crosslinguistically, along with a preference
for pronouns as arguments, may impact the observation that arguments
are shorter overall than predicates. This pattern aligns with the
processing claim that ``heavier'' words (or word classes in this
case) occur later in a sentence, partly validating Hypothesis 1 of
study 1. The counter-argument, also from the data, is that nouns are
longer than verbs in SV languages and shorter than verbs in VS languages
when frequency is considered, validating Hypothesis 2 of study 1.
This suggests that speakers prefer to place heavier (more ``contentful'',
or perhaps ``novel'') constituents before lighter ones crosslinguistically
in many cases.

These seemingly antithetical results partially validate both the ``efficiency''-based
theory and the ``surprisal'' theory, highlighting the interaction
of two pressures: one whereby speakers prioritize lighter constituents
(possibly for the purposes of production/processing efficiency), and
one where they prioritize heavier constituents (possibly for informational
purposes). The two perspectives may in fact be looking at the same
data from different angles. On the one hand, grouping all arguments
into a single class highlights the overall impact of shorter strings,
since the most frequently used arguments (pronouns, demonstratives)
are typically shorter crosslinguistically. On the other, more ``content-ful''
word types that aid planning and inference or predictability (nouns,
verbs) will typically be longer crosslinguistically.

These two pressures can be (and, in fact, have been) considered as
two different kinds of ``efficiency'': \emph{production} efficiency
vs \emph{communicative} efficiency. Here \emph{production efficiency}
is related to length of words (as per \citealt{Wasow:2002aa,Wasow:2022aa,Hawkins:2014aa})
- speakers use shorter words whenever they can to efficiently produce
utterances (see also \citealt{Mahowald:2013aa}). The counter-pressure,
\emph{communicative efficiency}, is related to making a speaker's
utterances more readily accessible/interpretable by recipients (as
per \citealt{Levshina:2021aa,Levshina:2023aa}) - speakers prioritize
words that contain more meaningful, informative content in order
to aid communication. Once context has been established, these longer
words can be replaced by more efficient (shorter) terms such as demonstratives
or pronouns.\footnote{For example, there is a common crosslinguistic pattern of using a
noun first to establish reference, then switching to pronouns, i.e.
\textquotedbl that tree, it is big, it will fall...\textquotedbl{}
See also discussion of the (lack of) encoding of subjects crosslinguistically
in \citet{Berdicevskis:2020aa}.}

I would argue, however, that conceptualizing ``efficiency'' in these
two distinct ways, as currently done in the literature, runs the risk
of confusing the two. It is in fact not efficient (from a processing
perspective) to maximize information, as this also tends to maximize
complexity. Instead, I propose a different framing of these principles.

\subsection{A \textquotedblleft Min-Max\textquotedblright{} theory of processing\label{subsec:A-Min-Max-theory}}

\begin{figure}[h]
\noindent \begin{centering}
\includegraphics[width=0.4\columnwidth]{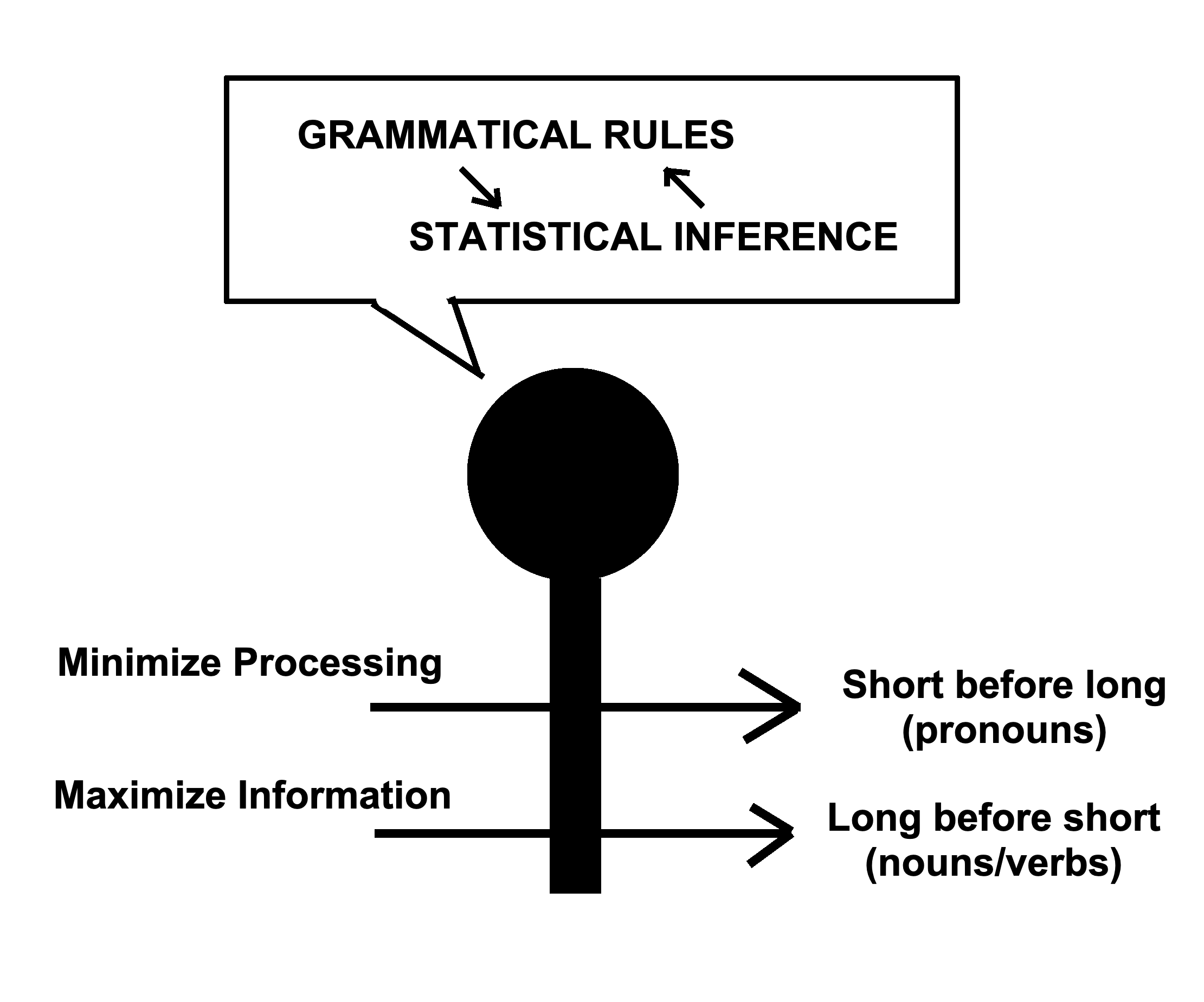}
\par\end{centering}
\caption{Min-Max schema\label{fig:Min-Max-schema-1}}
\end{figure}

The current proposal re-formulates or refines Hawkins' PGCH to incorporate
(increased) complexity as a motivation for the order of constituents,
bearing some resemblance to proposals in \citet{Levshina:2023aa}.
I describe this as (to borrow a term from computer gaming)\footnote{A ``min-maxer'' in gaming terminology is an individual who minimizes
their effort while maximizing their gains. This can be in terms of
focusing their efforts on activities in a game that give the best
returns, or by minimizing survival stats on a character in order to
maximize damage (as in a ``glass cannon'' build for ranged attackers
in an RPG).} the ``Min-Max'' theory of processing (schematized in Figure \ref{fig:Min-Max-schema-1}),
whereby speakers seek to minimize their processing effort while maximizing
the information in their utterances. These are competing/parallel
pressures that speakers are constantly balancing in production (or
``performance'') in order to choose the best strategy for their
immediate goals. As speakers produce variants along a cline of grammaticality,
these variants compete on the basis of multiple factors, including
how well such variants support Min-Max-ing in relation to other variants,
leading to language change.

These two competing pressures offer a more complete (and parsimonious)
explanation for word order change (and possibly other kinds of linguistic
change) than existing theories of language. In one possible scenario
the frequent use of short pronominal forms encourages drift toward
an argument-initial order, and nouns subsequently lengthen as they
provide more information to speakers and move toward an initial position.
This, combined with the preponderance of shorter pronominal forms
crosslinguistically, may provide an explanation for why the majority
of languages reported in the world are SV. Conversely, if a language
has or develops longer verbs, the preference for these heavier constituents
to occur first could put pressure on the language to maintain or develop
predicate-initial order, which would explain why some VS languages
are quite stable. In each case the direction of development would
depend on a number of factors that have been implicated in discussions
of language change and grammaticalization (see \citealt{Heine:2002fk,Matras:2011kx,Narrog:2011aa}).
However, the effect of word length is clearly a major part of the
explanation for why we can observe (and predict) such a difference
in word order between Ancient and Modern Hebrew as well as Classical
and Egyptian Arabic via their respective corpora.

The importance of frequency to any account of language evolution is
that frequency reflects usage. The most frequent items in a corpus
are those that are most used by speakers of a language. As such, under
a statistical learning paradigm, they are most likely to influence
generalizations made by speakers/hearers based on their input. This
means that any discussion of linguistic forms over time ought to consider
frequency as a factor in how such forms change.

As an example to illustrate the potential impact of this theory, I
offer the following thought experiment involving a hypothetical language.
This hypothetical language is limited to nouns and verbs, which the
speakers use in their interactions, in goal-directed behavior such
as making tools. Nouns and verbs are the same length, and the informational
content of each word is simply based on their communicative impact
in the interactions. As speakers use the nouns and verbs to communicate
about their tool-making, there is no pressure from noun/verb lengths
on order of constituents, with order being determined by the need
to identify objects/actions in immediate interaction. Over time, as
speakers innovate and additional information begins to get encoded
(such as deixis), the information will become associated with either
nouns or verbs, possibly resulting in additional morphology, and potential
affixation. The relevant noun/verb associated with such morphology
then becomes “heavier” and provides more information for the interaction.
This makes them more likely to be placed at the beginning of an utterance,
since that makes it easier to predict what follows. The information-rich
noun/verb has great utility and therefore has high frequency, becoming
notable and relevant to statistical inference by language learners,
who may then generalize an ordering rule that places the heavier constituent
at the front of the speech stream.

How does this relate to language change? In this thought experiment,
we could consider what might happen if a language that had developed
markers on verbs only (and therefore had VS order) innovated a new
set of tools for particular tasks, but with only a single word for
“tool”. As the tools became more prevalent, usage increased in interaction,
and each tool began to acquire affixes that identified their specific
use and other relations. This developed to the point that a listener
could infer, from the named object, a particular context and subsequent
utterances. At this point the nouns have become “heavier” than the
verbs, with the resulting stronger pressure to place them before verbs
in the speech stream, and the language would realize SV order.

While this is only a thought experiment, it does align with what we
know about the development of pidgins and creoles. Pidgins are languages
that arise out of contact situations and tend to lack morphology.
These can develop into creoles as they acquire native speakers (children
who grow up speaking the pidgin), and they begin to exhibit increased
morphological complexity as the younger speakers innovate forms. The
possibility laid out in this thought experiment is also supported
by the finding in this paper that the length of particular word classes
can predict the basic word order of a language. In turn, this leads
to some general predictions: if, on average, nouns are longer than
verbs in a language, it is highly likely to have SV word order. If,
on the other hand, verbs are longer, it is highly likely to have VS
word order. The effect of frequency highlights the importance of conducting
large-scale corpus based studies to investigate this and related concerns.
Since languages are used by speakers and such speakers use certain
words more than others, we can only observe this frequency effect
(or ``weight'') of word class length via corpora.

The existence of multiple competing pressures affecting the development
of language, as described by this Min-Max theory, is supported by
work in clinical psychology (e.g. \citealt{Rezaii:2023aa}) and language
modeling (e.g. \citealt{Hahn:2022aa}), as well as findings from information
theory (\citealt{Zaslavsky:2020aa,Tucker:2025aa}). In a study of
patients with nonfluent aphasia, \citet{Rezaii:2023aa} showed that
patients with this condition compensate for their deficit in producing
complex grammatical structures by increasing the information content
of their utterances, dropping high-frequency (function) words in favor
of low-frequency (content) words (i.e. ``maximizing information'').
By modeling variations in grammatical structure based on corpora from
80 languages, \citet{Hahn:2022aa} found that the resulting models
were less optimized for dependency length (DL) and information locality
(IL) than the original structures of each language, suggesting that
languages evolve to efficiently balance the two considerations of
DL and IL.\footnote{Notably, Hahn \& Xu propose a theory that the interaction between
Dependency Length (DL) and Information Locality (IL) accounts for
word order variation and change, but neither of the two concepts,
as expressed, make claims about directionality/order, only about the
relationships between (internal) constituents. In contrast, the Min-Max
theory makes the claim that speakers balance information content (as
reflected by longer nouns in SV languages and longer verbs in VS languages)
with processing/production concerns, and that this interaction is
what drives the evolution of word order. As nouns lengthen or verbs
shorten in a language, this will affect their placement in a stream,
which then impacts the overall linear order observed in surface structures.
That is, the Min-Max theory makes a universal claim regarding language-external
pressures, while DL and IL are language-internal concepts.} These and other studies indicate that language users are sensitive
to the pressures of minimizing processing and maximizing information
content.

The sensitivity of language users to processing efficiency and information
content has also been a major topic of investigation in information
theory, which is concerned in part with the encoding and decoding
of information streams. One information-theoretic proposal is that
semantic systems evolve following an ``Information Bottleneck''
principle (IB; \citealt{Tishby:1999aa}), whereby ``optimal representations
satisfy a tradeoff between compressing X, i.e. minimizing the representational
complexity, and maximizing the relevant information about Y'' (\citealt{Zaslavsky:2020aa}:
14). Zaslavsky shows that IB can account for the development of color
naming systems in language, and additional research shows the utility
of IB in machine learning and distributional models of semantics (\citealt{Slonim:2000aa}).

Implications of this interaction have most recently been examined
in agent-based emergent learning simulations with the addition of
a ``utility'' principle (\citealt{Tucker:2025aa}) – the interaction
between ``utility'', ``informativeness'', and ``complexity''
allows more human-like communication to emerge among agent models
in their study. Such a finding is very much supported by the results
reported here, though I would argue that the Min-Max theory offers
a more parsimonious account of human behavior as related to language
specifically. While the current corpus evidence applies at the word
level in relation to basic word order, it seems likely that the minimization
of processing difficulty would interact with the maximization of information
content at multiple (hierarchical) levels. Speakers min-max in order
to select words in a stream (``informativeness'' and ``complexity''
as per Tucker et al.), while simultaneously min-maxing to select relevant
structures related to their communicative goals (Tucker et al.'s ``utility'').

As noted above, the Min-Max theory also bears some resemblance to
Levshina's (\citeyear{Levshina:2023aa}) proposals, described in terms
of \emph{communicative efficiency} and \emph{accessibility}. However,
such indicators cannot fully explain certain patterns in grammar such
as ``harmonic'' word order. A recent proposed explanation for such
patterns involves frequency of word classes, whereby \citet{Mansfield:2025aa}
suggest that ``harmonic ordering is really a frequency effect: as
replication converges on a phrase structure with a consistent linear
order, the most frequent word class tends to be at one edge.'' Although
they did not test token length in their model, it is notable that
in Study 1 (§\ref{subsec:Study-1:-Word} above) frequency is a major
factor in identifying the relationship between length and word order,
suggesting that the most frequent word class in their model may also
be the longest element of the domain they are testing, which would
be driven by the maximization of information principle. The Min-Max
theory might also provide an explanation for non-parallel (hierarchical)
harmonic word order, which is not explained by their model.

The Min-Max theory may additionally explain why word forms remain
long (or lengthen), as described for particular languages (see \citealt{Cristofaro:2021aa,Haspelmath:2021aa,Petre:2017aa}).
Here, again, the maximization of information principle would support
the development or retention of particular (long) word forms, as speakers
balance this need against the minimization of processing difficulty.
As noted above, this proposal is supported by evidence from color-naming
systems, which show patterns in keeping with IB (\citealt{Zaslavsky:2020aa}).
The Min-Max theory would suggest that this principle does not only
hold for ``forms'' (of, i.e. color words), but also governs ``behavior'':
speakers behave in ways that are consistent with maximizing information
in streams (via word order), while minimizing processing effort.

Such a theory can be conceived of as a universal (underlying) pressure
that applies generally to how language is realized. From an innateness
perspective, speaker choices impacted by such pressures would lead
to a re-analysis or re-formation of rules that guide surface realizations.
From a usage-based perspective such pressures would lead to the development
of new surface variants. These are not necessarily opposing views
(see \citealt{Rastelli:2025aa}), and the outcomes are quite similarly
dependent on language being viewed as a complex adaptive system, whereby
diachronic changes may have a combination of both ``result-oriented''
and ``source-oriented'' explanations (see \citealt{Cristofaro:2019aa,Schmidtke-Bode:2019aa}).

Another observation from the results reported here is that information
content seems to trump length. We find that long nouns are used with
more frequency in SV languages, while long verbs are used with more
frequency in VS languages. This indicates that (maximizing) information
content is more important to speakers than (minimizing) processing
complexity. This could explain the ``agent/subject-first preference''
in experiments (\citealt{Culbertson:2012aa,Sauppe:2023aa}), as the
``information maximization'' pressure wins out in cases where this
most efficiently meets communicative goals.

Further, the data shows that all languages have a relatively high
proportion of sentences in which the argument occurs before the predicate.
In VS languages, the proportion of argument-initial to predicate-initial
sentences is 1/2 or more (min: 0.56 n/v, 0.65 a/p), highlighting that
even VS languages still place the noun/argument first in many constructions,
and suggesting that there is a strong efficiency pressure toward SV
order. Similarly, while all SV languages have a high proportion of
N1 sentences (max 2.43 n/v, 7.17 a/p), the number of V1 constructions
is non-zero, supporting a gradient approach to word order (\citealt{Levshina:2023ab}).

The idea that speakers balance multiple considerations in production
is not new (\citealt{Pike:1967lr,Jackendoff:2005yq,Serzant:2022aa,Horberg:2023aa})
but what is novel is the idea that all humans experience a specific
\emph{universal} pressure to balance production/processing efficiency
with information content. This pressure is observable crosslinguistically,
and would apply at multiple processing levels, regardless of whether
a speaker is accessing a universal grammar (``innate'') or inferring
structure on the basis of the input they receive (``usage''). Following
on from \citeauthor{Zipf:1949aa}'s and subsequent observations, there
is no doubt that the length of strings is an important factor in processing
streams, but clearly not all strings are created or treated equally
by speakers. The studies presented here show that nouns are treated
differently (or at least have a different distributional relationship)
than other kinds of arguments, which may then lead to different realizations
of word order crosslinguistically. The pressures identified in this
paper have wide-ranging implications and applications for any conceptualization
of how language operates, whether in terms of understanding the intricacies
of individual linguistic scenarios, designing intelligent systems
that use language, or developing clinical interventions to assist
speakers in various ways.

\subsection{Limitations and future directions\label{subsec:Limitations}}

One limitation in this set of studies is the use of automated processes
to identify part of speech tags in the data. Ideally, all the data
should be hand-annotated for parts of speech, which would allow for
better comparison of the different word categories across languages.
Given the time consuming nature of this task and the need for expert
annotators, this is not yet possible, but ongoing research could address
this by sourcing experts to annotate the data. Although I have shown
(in \citealt{Ring:2026ab} and via the linked repository) that the
translation/alignment method does correspond well to hand-tagged data,
it may be that this could be improved by tagging a small amount of
data for each language in the PBC in order to train POS taggers that
perform better than the current translation method.

Another limitation is that the dataset used, while exceedingly large
in comparison to previous efforts, still only represents about a quarter
of the world's languages. Since the majority of the world's languages
are not represented here, it is not clear that these findings can
be fully generalized to all languages of the world, though I have
shown in section \ref{subsec:Study-3:-Testing} that much can be learned
from sampling. Additionally, the contrast between intransitive SV/VS/free
word order is somewhat of an oversimplification of word order patterns.
Given the current obstacles to annotation, it is a necessary simplification
for this study, however future work could expand to a more diverse
set of word/constituent orders and potential relationships with word
length and/or complexity. Given the nature of the dataset used, it
is also possible to pursue a gradient approach to word order, as advocated
by \citet{Levshina:2023ab}. The issue of word complexity could additionally
use further expansion to cover a broader range of possibilities than
simply the length of a word based on a set of (roman) characters.

A final note is that our conclusions are only as good as the data
that we have available to us. The \emph{taggedPBC} is a significant
step in the right direction, toward a representative sample of large
tagged corpora for languages of the world. Systematic investigation
of this dataset can give us insight into language typology (the current
state of language), as well as pressures that impact this realization
in diachrony, both in terms of how languages have developed (their
history) and how they might develop (their future). On the basis of
this investigation I have proposed a parsimonious ``Min-Max'' theory
of processing and information structure that underlies language evolution,
whereby speakers minimize processing effort while maximizing information
gain. This proposal is in keeping with at least one existing information-theoretic
proposal (IB) as well as theories of processing (efficiency; PGCH;
surprisal) and can explain the correlation between the length of nouns/verbs
and basic word order, given that these features allow us to predict
the word order of a language. It is as yet unclear whether such a
principle would apply at multiple hierarchical levels within language,
which is an area that deserves further careful investigation with
datasets such as the \emph{taggedPBC}. For now, I have demonstrated
that noun/verb lengths are more explanatory than descent or geography
and that for languages with corpus data, word length predicts word
order.

\bibliographystyle{linguistics}
\phantomsection\addcontentsline{toc}{section}{\refname}\bibliography{my-bibliog}

\end{document}